\documentclass{article}

\usepackage[nonatbib,preprint]{neurips_2023}

\usepackage[utf8]{inputenc} 
\usepackage[T1]{fontenc}    
\usepackage{hyperref}       
\usepackage{url}            
\usepackage{booktabs}       
\usepackage{amsfonts}       
\usepackage{nicefrac}       
\usepackage{microtype}      
\usepackage{xcolor}         

\usepackage{microtype}
\usepackage{graphicx}
\usepackage{subfigure}
\usepackage{booktabs} 
\usepackage[linesnumbered,ruled]{algorithm2e}
\usepackage{float}
\usepackage{algorithmic}
\usepackage{bm}
\usepackage{multirow}
\usepackage{microtype}
\usepackage{enumitem}
\usepackage{nicefrac}
\usepackage{soul}
\usepackage{amssymb}
\usepackage{pifont}
\newcommand{\cmark}{\textcolor{green}{\ding{51}}}%
\newcommand{\xmark}{\textcolor{red}{\ding{55}}}%
\usepackage{colortbl}
\usepackage{comment}
\usepackage{soul}

\usepackage{hyperref}
\usepackage{wrapfig}
\usepackage{caption}

\usepackage{amsmath}
\usepackage{amssymb}
\usepackage{mathtools}
\usepackage{amsthm}

\newcommand\norm[1]{\lVert#1\rVert}

\usepackage[capitalize,noabbrev]{cleveref}

\theoremstyle{plain}
\newtheorem{theorem}{Theorem}[section]
\newtheorem{proposition}[theorem]{Proposition}
\newtheorem{lemma}[theorem]{Lemma}

\theoremstyle{definition}

\theoremstyle{remark}
\newtheorem{remark}[theorem]{Remark}

\usepackage[textsize=tiny]{todonotes}

\title{Securing Deep Generative Models with Universal Adversarial Signature}

\author{{Yu Zeng$^*$,
\hfill
Mo Zhou\thanks{Equal contribution},
\hfill
Yuan Xue,
\hfill
Vishal M. Patel
}\\
{\tt\small \{yzeng22,mzhou32,yuanxue,vpatel36\}@jhu.edu}\\
{Johns Hopkins University}\\
}

\begin{document}

\maketitle

\begin{abstract}
Recent advances in deep generative models have led to the development of methods capable of synthesizing high-quality, realistic images.  These models pose threats to society due to their potential misuse. Prior research attempted to mitigate these threats by detecting generated images, but the varying traces left by different generative models make it challenging to create a universal detector capable of generalizing to new, unseen generative models. In this paper, we propose to inject a universal adversarial signature into an arbitrary pre-trained generative model, 
in order to make its generated contents more detectable and traceable. First, the imperceptible optimal signature for each image can be found by a signature injector through adversarial training. Subsequently, the signature can be incorporated into an arbitrary generator by fine-tuning it with the images processed by the signature injector. In this way, the detector corresponding to the signature can be reused for any fine-tuned generator for tracking the generator identity. The proposed method is validated on the FFHQ and ImageNet datasets with various state-of-the-art generative models, consistently showing a promising detection rate. Code will be made publicly available at \url{https://github.com/zengxianyu/genwm}.
\end{abstract}

\section{Introduction}
\label{sec:1}

Recent advances in deep generative models~\cite{gan-survey,ddm-survey} have enabled the generation of highly realistic synthetic images, which benefits a wide range of applications such as neural rendering~\cite{image-to-image-translation,park2019semantic,chan2022efficient,li2020bachgan}, text-to-image generation~\cite{dalle2,reed2016generative,zhang2017stackgan,saharia2022photorealistic}, image inpainting~\cite{feat-learning-inpainting,yu2018generative}, super-resolution~\cite{srgan}, among others. As a side effect, synthetic but photo-realistic images pose significant societal threats due to their potential misuse, including copyright infringement, the dissemination of misinformation, and the compromise of biometric security systems. 

\begin{figure}
\begin{center}
\includegraphics[width=0.9\textwidth]{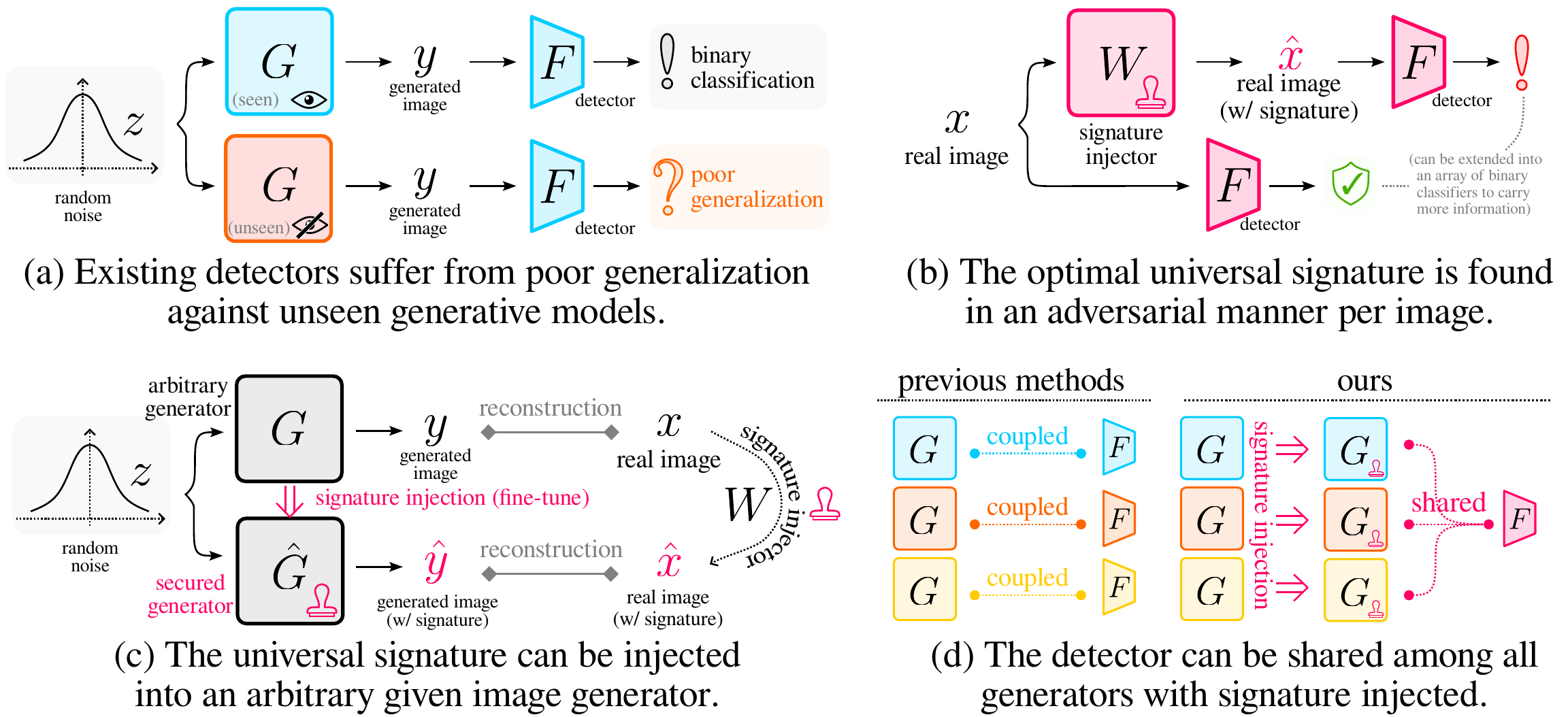}
\end{center}
\caption{Illustration of securing deep generative models through universal adversarial signature. 
%
%
%
%
}
\label{fig:1}
\vspace{-1em}
\end{figure}

To mitigate these risks, one straightforward approach is to imprint digital watermarks on generated images during a post-processing phase. However, this post-processing step is usually decoupled from the main model, making it difficult to enforce. Therefore, recent work has focused on a more enforceable solution: using a deep model as a detector to identify synthetic images~\cite{deepfake-face,liu2021spatial,deepfake-conv-trace,wang2022m2tr,sha2022fake}. They manifest effectiveness against known generators, \emph{i.e.}, 
those involved in the training dataset, but suffer from a performance drop against unseen generators. This is due to the variability of the model ``signatures'' hidden in the generated images across different models, as illustrated in Fig.~\ref{fig:1} (a). 
Consequently, these detection-based systems require frequent retraining upon the release of each new generator to maintain their effectiveness, which is impractical in real-world applications. 

In this work, we propose a more robust approach to identify synthetic images by integrating a model-agnostic ``signature" into any pre-trained generator. Since the signature is concealed within the model parameters, it becomes non-trivial for malicious users to erase, and is inevitably included in the generated contents, thereby facilitating detection. By using a universal signature (\emph{i.e.}, model-agnostic signature), we can leverage the same detector to identify images from different generators, eliminating the need for retraining the detector with the introduction of new generators. 

To determine the optimal signature for images, 
we first train a signature injector $W$ in an adversarial manner
against a classifier $F$ (the detector). In particular, the injector $W$ learns to add a minimal alternation $\kappa$ 
to a given image $\bm{x}$ to produce a slightly modified image $\hat{\bm{x}}$. The injector aims to make the alternation $\kappa$ as small as possible in order to retain image quality, 
while simultaneously maximizing the detector's accuracy, as shown in
Fig.~\ref{fig:1} (b). Importantly, the detector $F$ is not necessarily designed to be a binary classifier. 
It can be a multi-class classifier that produces a multi-bit binary code to convey additional information to help track and identify the source of a generated image. 

To implant such signatures into an arbitrary pre-trained image generative model $G$,
we fine-tune $G$ using a set of images processed by $W$, resulting in a secured generator $\hat{G}$, as demonstrated in Fig.~\ref{fig:1} (c). 
Images generated by the secured generator $\hat{G}$ can be identified by $F$ as $\hat{G}$ inherits the signatures from $W$. 
In addition, as shown in Fig.~\ref{fig:1} (d), the detector $F$ is no longer associated with specific generators during the training process, and therefore can be shared across different generators. As the injector $W$ and detector $F$ can be reused for different pre-trained generators, the adversarially learned signature becomes universal (model-agnostic) among all secured generators.


%

To demonstrate the effectiveness of the proposed method, we conduct extensive experiments on the FFHQ~\cite{stylegan} and ImageNet~\cite{imagenet} datasets. 
Three state-of-the-art generative models, namely 
LDM~\cite{ldm},
ADM~\cite{ADM},
and StyleGAN2~\cite{stylegan2}, 
are used in evaluations.
The experimental results demonstrate that a given generative model can learn to add the adversarial signatures to its outputs, making them more recognizable by the generated image classifier, while not sacrificing the image quality.

%
%
The contributions of this paper are summarized as follows:
\begin{enumerate}[noitemsep,nosep]
%
\item We propose to learn the optimal universal signatures through adversarial learning against a classifier (\textit{i.e.} the detector). 
\item We propose to inject the universal adversarial signatures to secure an arbitrary pre-trained image generative model. 
Secured generative models can share the same detector. 
%
\item Our proposed universal adversarial signature is capable of carrying additional information, such as the generator identity, for tracking the source of the images.
\end{enumerate}

\section{Related Works}
\label{sec:2}

\noindent\textbf{Deep Generative Models}%
~\cite{gan-survey,ddm-survey} have been greatly improved recently, enabling realistic large-scale image and text synthesis~\cite{dalle2},~\cite{gpt3}.
This field has undergone various mainstream models, including autoregressive models~\cite{PixelCNN++}, variational autoencoders~\cite{huang2018introvae}, normalizing flows~\cite{kingma2018glow}, generative adversarial models (GANs)~\cite{gan,stylegan2,vqgan}, and more recently, denoising diffusion models (DDMs)~\cite{ddpm,ldm}. 
In particular, GANs and DDMs are capable of 
imposing threats to society due to their potential abuse.
This paper focuses on mitigating their potential threats. 

\noindent\textbf{Generated Image Detection}
is committed to mitigating the potential threats of the generated images. %
%
The existing methods
extract features to discover artifacts either in the spatial domain~\cite{deepfake-face,deepfake-conv-trace,sha2022fake} or frequency domain~\cite{liu2021spatial,wang2022m2tr}. However, these passive detection models 
may not generalize well for unseen 
generative models.
In this paper, learning the adversarial signature entails actively modifying a generator to make its outputs more recognizable, which is different from the existing work focused on generated image detection. 
%
The scope of this paper is general-purpose generated image detection, which is not limited to a specific type of media such as deepfake. 


\noindent\textbf{Image Watermarking}%
~\cite{cox2007digital,begum2020digital} can limit the misuse of digital images, including the generated ones. 
Although watermarks vary in their visibility~\cite{nikolaidis1998robust,zhou2018robust}, it is difficult for them
to achieve robustness, imperceptibility, and high capacity at the same time~\cite{begum2020digital}.
Besides, deep-learning-based methods involve adding templates to the real images~\cite{asnani2022proactive}, or inserting watermarks into the generated image~\cite{zhao2023proactive}. 
However, these methods are subject to an impractical assumtion that malicious users will apply
watermarks. 
%
Instead, we modify generative models to make adversarial signatures inevitable.


\noindent\textbf{Neural Network Fingerprinting} 
addresses the challenges of visual forensics and intellectual property protection posed by deep models~\cite{peng2022fingerprinting,liu2021watermarking}. 
%
%
Model-specific GAN fingerprints, either learned~\cite{yu2019attributing} or manually-crafted~\cite{gan-finger-craft}, can be used for generated image detection and source tracking,
but still has to be re-trained against new generators. 
In contrast, our detector can be reused for any future generator. 

\section{Our Approach}
\label{sec:3}

Given a set of images $X$, a typical deep generative model $G$ learns the underlying distribution $p_X$ of $X$, and can generate a realistic image from a random noise $\bm{z}$, \emph{i.e.}, $\bm{y}\triangleq G(\bm{z})$.
Due to the threats posed by the potential abuse of the outputs from the generator $G$, it is necessary to develop a classifier $F$ to distinguish the generated (signed) images from real ones, where $F(\cdot)\in (0,1)$. 
A real image $\bm{x}\in X$ is labeled as $0$, while a generated image $\bm{y}$ is labeled as $1$. 


As discussed in Section~\ref{sec:1}, we explore modifying the parameters of a given generator $G$, to make its outputs more recognizable by $F$, 
and hence securing the generator $G$. 
%
%
%
Our approach is a two-stage method.
Firstly, a signature injector $W$ learns to generate adversarial signatures, while a classifier $F$ learns to detect them.
The signature injector $W$ is subsequently used for teaching an arbitrary generative model $G$ to generate recognizable images. 
The proposed method is illustrated in Figure~\ref{fig:1} and summarized in Algorithms~\ref{alg:1}-\ref{alg:2}.

\subsection{Optimal Adversarial Signature}
\label{sec:31}
%

Consider a system consisting of a signature injector $W$ and a classifier $F$. 
%
In the optimal case, $F$ can discriminate the signed images from clean images based on the subtle and imperceptible alternation made by $W$ (imperceptibility). The system is robust to image restoration attack if augmented by noise (persistency),~\textit{i.e.} the signature cannot be removed by an image restoration model $M$. The following propositions state the imperceptibility and persistency of the adversarial signatures in details. 
%
%
\begin{proposition}
\label{prop1}
(\textit{\textbf{Imperceptibility}}) There exist optimal pairs of signature injector $W$ and classifier $F{:}~ \mathbb{R}^n {\mapsto} \{0,1\}$, so that for any image $\forall \bm{x} {\in} \mathbb{R}^n$, $\forall \epsilon {>}0$,
its distance to the signed image $W(\bm{x})$
is smaller than $\epsilon$, 
and $F$ correctly discriminates them, 
\emph{i.e.},
$\norm{W(\bm{x}){-}\bm{x}}{<}\epsilon$, and
$F(W(\bm{x}))\ne F(\bm{x})$. 
\end{proposition}

\begin{proposition}
\label{prop2}
Let $\bm{e}$ be a zero-mean, positive-variance additive noise. There exist noise augmented $W, F$ that satisfy the following condition: $\forall \epsilon > 0, \mathbb{E}_{\bm{e}}[\norm{W(\bm{x}+\bm{e})-\bm{x}}]< \epsilon$ and $F(W(\bm{x}))\ne F(\bm{x})$. 
\end{proposition}

\begin{proposition}
\label{prop3}
(\textit{\textbf{Persistency}}) The noise augmented
$W, F$ stated in Proposition~\ref{prop2} is robust to image restoration attack, as optimizing $\min_M \mathbb{E}_{\bm{x},\bm{e}}[ \norm{ M(W(\bm{x}+\bm{e})) - \bm{x} }]$ will result in $M$ being an identity mapping. 
\end{proposition} 
\noindent \textit{Proof. }Please refer to the supplementary material. 
\begin{remark}
Intuitively, when $W(\bm{x}+\bm{e})$ is close enough to $\bm{x}$, the training of $M$ to remove signatures tends to fall into a trivial sub-optimal solution of copying the input to the output. 
Therefore, even if $W$ is disclosed to malicious users,
it is still difficult to erase the signature.
\end{remark}

\subsection{Universal Adversarial Signature Injector}
\label{sec:32}

Given an image $\bm{x}$, the signature injector model $W$ adds an imperceptible alternation to it, resulting in a ``signed'' image $\hat{\bm{x}}\triangleq W(\bm{x})$, of the same size as $\bm{x}$. 
The difference $\kappa \triangleq \hat{\bm{x}}-\bm{x}$ is termed as the ``adversarial signature'', which varies with the input image $\bm{x}$ and the injector $W$. 
Meanwhile, the classifier $F$ aims to discriminate the signed image $\hat{\bm{x}}$ from the clean image $\bm{x}$. 
In this paper, the signed image $\hat{\bm{x}}$ is labeled as $1$, while the clean image $\bm{x}$ is labeled as $0$. 

To find the desired pair of $W$ and $F$ as discussed above, the goal is to ensure that
the signed image $\hat{\bm{x}}$ is as close to the clean image $\bm{x}$ as possible, while the classifier $F$ should correctly recognize the signed images.
The goal can be expressed as the following optimization problem:
\begin{equation}
\min_{W,F} \   \mathbb{E}_{\bm{x}}  \norm{ W(\bm{x})-\bm{x} }_2^2,  
\;\;\;\mbox{s.t.}\;\;\;~ \mathbb{E}_{\bm{x}} \left[ F(\bm{x})+(1-F(W(\bm{x})) \right]= 0. 
\end{equation}
By introducing the Lagrange multiplier, we obtain the following loss function:
\begin{equation}
\label{eqn_loss0}
\mathcal{L} = \mathbb{E}_{\bm{x}} [ \underbrace{\norm{W(\bm{x})-\bm{x}}_2^2}_{L_\text{rec}} + \lambda \underbrace{(F(\bm{x}) + 1-F(W(\bm{x})))}_{L_\text{cls}} ]. 
\end{equation}
The $L_\text{rec}$ term in Eq.~\eqref{eqn_loss0} is the mean squared error that enforces the signatures to be imperceptible (not obviously impacting the image quality). The $L_\text{cls}$ term can be seen as a classification loss that encourages the classifier to distinguish the signed images from the clean images. 

In practice, we find directly optimizing Eq.~\eqref{eqn_loss0} through gradient descent methods results in $\lambda=0$, and
the model 
copying the input to the output. Therefore, we empirically fix $\lambda$ to a small value. In addition, we replace the $L_\text{cls}$ part with the commonly used cross-entropy loss.
Therefore, $W$ and $F$ are jointly trained by optimizing the following approximated loss function:
\begin{align}
&L = \mathbb{E}_{\bm{x}\sim p_X} \{
L_\text{rec}(\bm{x}; W) + \lambda \cdot L_\text{cls}(\bm{x}; W, F) \}\label{eq:advwm}, \\ \text{where} \;\;\;
&L_\text{rec}(\bm{x};W)=\norm{W(\bm{x})-\bm{x}}_2^2, \label{eq:rec} \\ \text{and} \;\;\;
&L_\text{cls}(\bm{x};W,F)=\log F(\bm{x}) + \log (1 - F(W(\bm{x}))). \label{eq:cls}
\end{align}
During training, the signature injector $W$ and the generated image classifier $F$ are, in fact, adversarial against each other. 
The minimization of $L_\text{cls}$ requires the injector $W$ to add a sufficiently large and easy-to-identify signature $\kappa$ to make $\hat{\bm{x}}$ separatable from $\bm{x}$; 
while the minimization of $L_\text{rec}$ requires the signature injector $W$ to shrink the norm of $\kappa$ for the sake of its imperceptibility, which makes the signed image $\hat{\bm{x}}$ more difficult to be separated from $\bm{x}$. 
The overall process of this stage is summarized in Algorithm~\ref{alg:1}.
Note, to make the signature $\kappa$ robust,
both the original image $\bm{x}$ and the signed image $\hat{\bm{x}}$ are transformed before being fed to $F$.
The transformations involve commonly used augmentation operations, which are detailed in Section~\ref{sec:4}.
%

Our method resembles letting $W$ produce adversarial examples to flip the prediction of $F$.
Albeit the goal is similar to the C\&W attack~\cite{cw-attack}.  Our method generates the signed images in a single forward pass (instead of iteratively), and jointly trains $F$ (instead of fixing its parameters ).
%


\begin{figure}[t]
\begin{minipage}[t]{0.49\linewidth}
\begin{algorithm}[H]
\caption{Training Signature Injector}\label{alg:1}
\begin{algorithmic}[1]
\STATE {\bfseries Input:} A set of images $X$; 
\STATE {\bfseries Output:} (1) Signature injector $W$;
\STATE ~~~~~~~~~~~~~~ (2) Binary classifier $F$;
\STATE Randomly initialize $W$ and $F$;
\FOR {$i=1$ {\bfseries to} MaxIteration\_stage1}
    \STATE Randomly sample $\bm{x} \in X$;
    \STATE $\hat{\bm{x}} \leftarrow W(\bm{x})$;
    \STATE $L_\text{rec} \leftarrow \| \hat{\bm{x}} - \bm{x} \|_2^2 $; 
    \STATE Random transformation for $\bm{x}$ and $\hat{\bm{x}}$;
    \STATE $L_\text{cls} \leftarrow \log F(\bm{x}) + \log (1-F(\hat{\bm{x}}))$; 
    \STATE $L \leftarrow L_\text{rec} + \lambda \cdot L_\text{cls}$;
    \STATE $\Delta W, \Delta F \leftarrow \nicefrac{\partial L}{\partial W}, \nicefrac{\partial L}{\partial F}$; 
    \STATE $W,F\leftarrow \text{Adam}(W,F; \Delta W, \Delta F)$;
\ENDFOR
\end{algorithmic}
\end{algorithm}
\end{minipage}
\hfill
\begin{minipage}[t]{0.49\textwidth}
\begin{algorithm}[H]
\caption{Securing an Image Generator}\label{alg:2}
\begin{algorithmic}[1]
\STATE {\bfseries Input:} A set of clean images $X$;
\STATE ~~~~~~~~~~~ A pre-trained generator $G$; 
\STATE ~~~~~~~~~~~ The signature injector $W$; 
\STATE {\bfseries Output:} A fine-tuned generator $\hat{G}$;
\STATE $\hat{G} \leftarrow G$; 
\STATE $\hat{X} \leftarrow \{ W(\bm{x}) | x \in X \}$; 
\FOR {$i=1$ {\bfseries to} MaxIteration\_stage2}
    \STATE Randomly sample a noise $\bm{z}$;
    \STATE Randomly sample $\hat{\bm{x}} \in \hat{X}$;
    \STATE Update $\hat{G}$ using $\hat{G}(\bm{z})$ and $\hat{\bm{x}}$;
\ENDFOR
\end{algorithmic}
\end{algorithm}
\end{minipage}
\vspace{-1em}
\end{figure}

\noindent\textbf{Binary Code Extension.} 
By extending the binary classifier $F$ to multiple outputs,
the adversarial signature will be able to carry additional information such as generator identity for tracking the source of the generated images.
To achieve this, we can first determine the binary code length as $n$ bits, which directly decides the number of all possible binary codes as $2^n$.
The selection of $n$ ($n>0$) depends on the number of user-predefined messages to be represented by the binary codes.
For instance, when $n=2$, the binary codes for generators are
\verb|01|, \verb|10|, and \verb|11|, while the code
\verb|00| is reserved for real images.
During the training process, a random binary code except for \verb|00| from the $2^n-1$ possible binary codes is chosen for every generated image.
Next, the single classification layer in $F$ is extended into $n$ classification layers in parallel for binary code prediction.
%
%
%

Meanwhile, the binary code is converted by a code embedding module into an embedding vector.
It comprises of two linear layers and SiLU~\cite{gelu} activation.
The resulting binary code embedding is fed into 
 injector $W$ via AdaIN~\cite{stylegan} after every convolution layer for modulating the signatures.
%
%
%
%
Note, in the default case where $n=1$, a constant vector is used as the binary code embedding.

\subsection{Securing Arbitrary Generative Model}
\label{sec:33}

In order to make the adversarial signatures inevitable, it would be better if they could be integrated into the model parameters through, for example, fine-tuning.
In this way, the outputs from the generators will be detectable by $F$, and hence the generative model is secured.
Therefore, in this stage, the signature injector $W$ will process the training data, based on which an arbitrary given (pre-trained) generative model is fine-tuned to learn the adversarial signatures.
This conceptually shifts the distribution the generator has learned towards the distribution of the signed images.


Specifically, given a set of training images $X$, the already trained signature injector $W$ is used to apply an adversarial signature to each image $\bm{x} \in X$, resulting in a signed image $\hat{\bm{x}}$.
Assume we have an arbitrary already trained deep generative model $G$, which can generate an image $\bm{y}$ from a random noise $\bm{z}$, \emph{i.e.}, $\bm{y}=G(\bm{z})$.
Then, the model $G$ is fine-tuned using the signed images, resulting in the model $\hat{G}$, which generates a signed image $\hat{\hm{y}}$ from a random noise $\bm{z}$, \emph{i.e.}, $\hat{\bm{y}}=\hat{G}(\bm{z})$.
By default, the concrete loss function during fine-tuning is consistent with the original training loss of $G$.
An optional loss term, \emph{i.e.}, $\xi \cdot \log (1 - F(\hat{G}(\bm{z})))$ can be appended to guide the training of $\hat{G}$ using the trained classifier $F$ (fixed), where $\xi$ is a constant that controls the weight of this loss term. 
The overall procedure of stage two is summarized in Algorithm~\ref{alg:2}.

As the fine-tuning process is agnostic to generator architecture, it is applicable to a wide range of generative models, including but not limited to GANs~\cite{gan-survey} and DDMs~\cite{ddm-survey}.
As the $W$ and $F$ are fixed in the second stage, they are reusable for different generators.

\textbf{Binary Code Extension.}
In this stage, a binary code can be assigned to a specific $G$.
Every signed image $\hat{\bm{x}}$ for fine-tuning $G$ is generated by $W$ with the assigned code.

\noindent\textbf{Inference Stage.}
As the fine-tuned model $\hat{G}$ is expected to learn the signatures, the classifier $F$ from the first stage can be directly used to identify whether $\hat{\bm{y}}$ is a generated (signed) image.
%
%
%
%
%
%
%

\section{Experiments}
\label{sec:4}

In this section, we present experimental results to demonstrate the effectiveness of the proposed method.
%
%
%
%
%
%
Our method is implemented in PyTorch~\cite{pytorch}. 
The code will be released in the future. 

\noindent\textbf{Datasets \& Models.}
We adopt the U-Net~\cite{ddpm} architecture for signature injector $W$, and ResNet-34~\cite{resnet} as the classifier $F$.
The proposed method is evaluated with two datasets: FFHQ~\cite{stylegan} and ImageNet~\cite{imagenet};
using three generative models:
%
LDM~\cite{ldm},
ADM~\cite{ADM},
and StyleGAN2~\cite{stylegan2} at $256\times 256$ resolution.
%
%
We use their official training code for the experiments, except for StyleGAN2.    
A third-party implementation\footnote{\scriptsize\url{https://github.com/rosinality/stylegan2-pytorch}}
 is used for StyleGAN2.
 We sample 1,000 images from FFHQ as the test set and use the remaining images for training. For experiments on ImageNet, we use the official training split for training, and sample 1,000 images from the validation split as our test set. The image quality (FID, PSNR) and classification accuracy (Acc, ROC) are evaluated on the test sets (1,000 images). The only exceptions are the FID scores in Tab.~\ref{tab:stage-2}, which are evaluated on 50K randomly generated images against the corresponding training sets following~\cite{ldm}. 

\noindent\textbf{Hyper-Parameters.}
In stage one,
the balance factor $\lambda$ in Eq.~\eqref{eq:advwm} is set as $0.05$ for the FFHQ dataset, and $1.0$ for the ImageNet dataset.
The batch size is set as $24$.
The models are trained using the Adam~\cite{adam} optimizer for $10^6$ iterations, with the learning rate of $10^{-4}$.
%
%
In stage two, we follow the parameter settings of the respective generative models.
The parameter $\xi$ is empirically set as $0.05$ for StyleGAN2, and $0$ for the remaining models.

\noindent\textbf{Data Augmentation.}
The image transformation operations used to process $\bm{x}$ and $\hat{\bm{x}}$ for training $F$ are random rotation (the angle is uniformly sampled within $[-30^\circ, 30^\circ]$), random horizontal flip (with $0.5$ probability), and Gaussian blur (the variance is uniformly sampled within $[0.01,10]$).
%
%
Any output of $W$ and input to $F$ will be clipped to $[0,1]$, and padded with the smallest constant error to make it an integer multiple of $\nicefrac{1}{255}$, to ensure validity as an image.

\noindent\textbf{Binary Code.}
By default, the binary code length is $n=1$, which means $F$ only predicts whether the input is generated or not.
%
%
For the $n>1$ case, we specifically choose $n=2$ to ensure a certain level of generator diversity, while avoiding some unnecessary experiment cost for demonstration.

\noindent\textbf{Evaluation Protocol.}
The experimental results are reported based on the test sets.
(1) Signature injector $W$: The $\kappa$ is expected to be imperceptible to retain the image quality of $\hat{\bm{x}}$ compared to $\bm{x}$. 
Therefore, $W$ is quantitatively evaluated using the peak signal-to-noise ratio (PSNR) and FID~\cite{FID} of its outputs.
(2) Generated image classifier $F$: The generated/real binary classification and binary code prediction are evaluated in classification accuracy.
%
%
%
(3) Generator $\hat{G}$: the fine-tuning process of $\hat{G}$ is expected to make $\hat{G}$ add adversarial signatures while retaining image quality.
Hence, the FID of the generated signed image $\hat{\bm{y}}$ and the accuracy of $F$ against $\hat{G}$'s outputs is reported.

\subsection{Validating $W$ and $F$ in the First Stage}
\label{sec:41}

\begin{table}[t]
\begin{minipage}[t]{0.49\textwidth}
\captionof{table}{Validating $W$ and $F$ in the first stage when
the length of the binary code is $n=1$.
The symbols ``$\uparrow$'' and ``$\downarrow$'' mean ``the higher the better'' and ``the lower the better'', respectively.
%
}
\label{tab:stage-1}
\begin{center}
 \resizebox{1.0\columnwidth}{!}{%
\setlength{\tabcolsep}{7pt}
\begin{tabular}{c|cc|c}
\toprule
\multirow{2}{*}{\bf Dataset} & \multicolumn{2}{c|}{Signature Injector $W$} & Classifier $F$\tabularnewline
 & PSNR $\uparrow$ & FID $\downarrow$ & Accuracy (\%) $\uparrow$\tabularnewline
\midrule
FFHQ & 51.4 & 0.52 & 100.0\tabularnewline
ImageNet & 38.4 & 5.71 & 99.9\tabularnewline
\bottomrule
\end{tabular}
 }
\end{center}
\vspace{-1.5em}
\end{minipage}
\hfill
\begin{minipage}[t]{0.49\textwidth}
\captionof{table}{Validating $W$ and $F$ in the first stage when the length of the binary code is $n=2$.
The symbols ``$\uparrow$'' and ``$\downarrow$'' denote ``the higher the better'' and ``the lower the better'', respectively.
}
\label{tab:stage-1-n}
\begin{center}
 \resizebox{1.0\columnwidth}{!}{%
\setlength{\tabcolsep}{7pt}
\begin{tabular}{c|cc|c}
\toprule
\multirow{2}{*}{\bf Dataset} & \multicolumn{2}{c|}{Signature Injector $W$} & Classifier $F$\tabularnewline
 & PSNR $\uparrow$ & FID $\downarrow$ & Accuracy (\%) $\uparrow$\tabularnewline
\midrule
FFHQ &44.9 &2.68 &99.9 \tabularnewline
\bottomrule
\end{tabular}
 }
\end{center}
\end{minipage}
\end{table}

\begin{wrapfigure}{r}{0.5\textwidth}
\centering
\vspace{-4em}
\begin{minipage}{0.24\columnwidth}
\includegraphics[width=\columnwidth]{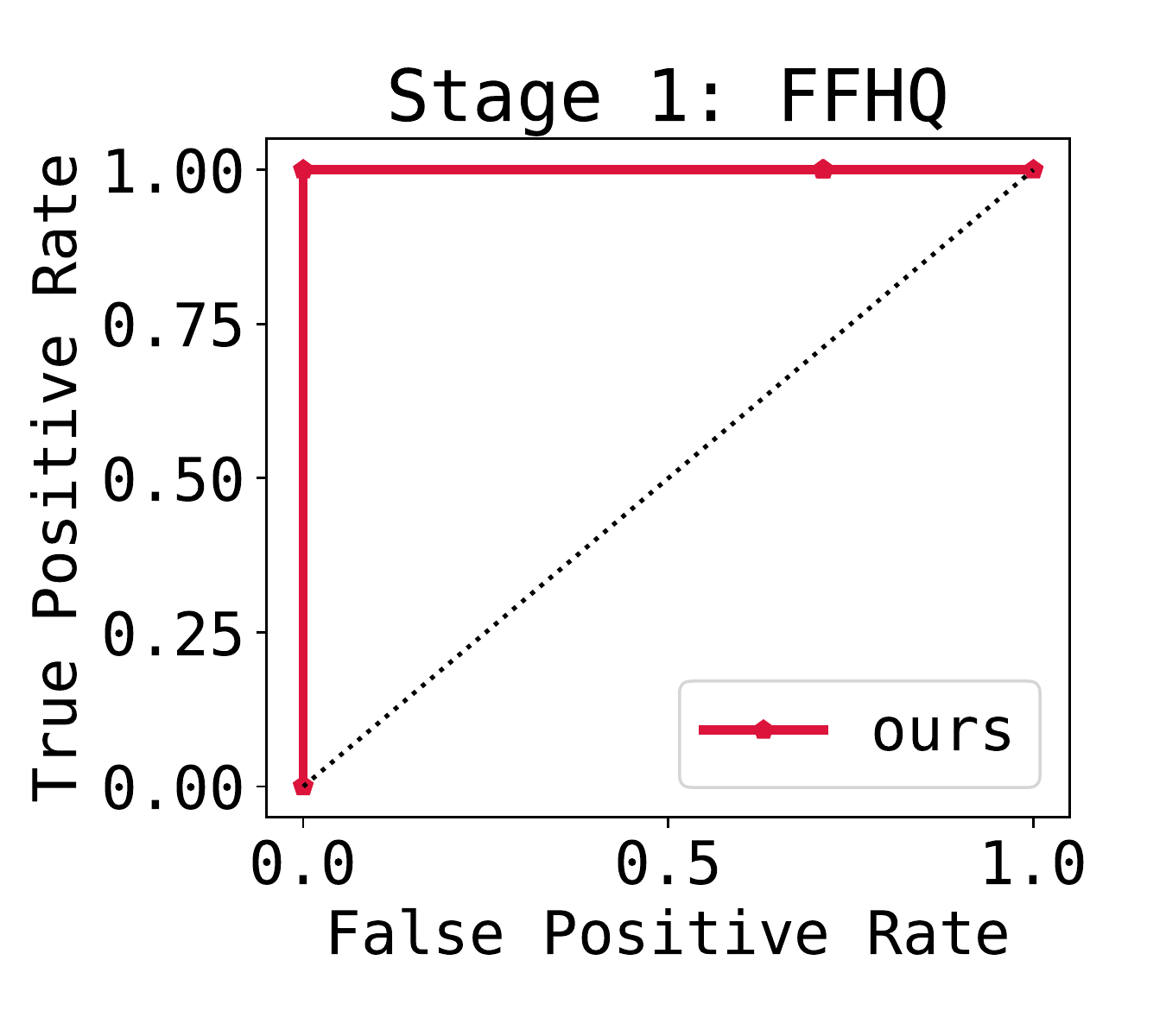}
\end{minipage}
%
%
\begin{minipage}{0.24\columnwidth}
\includegraphics[width=\columnwidth]
{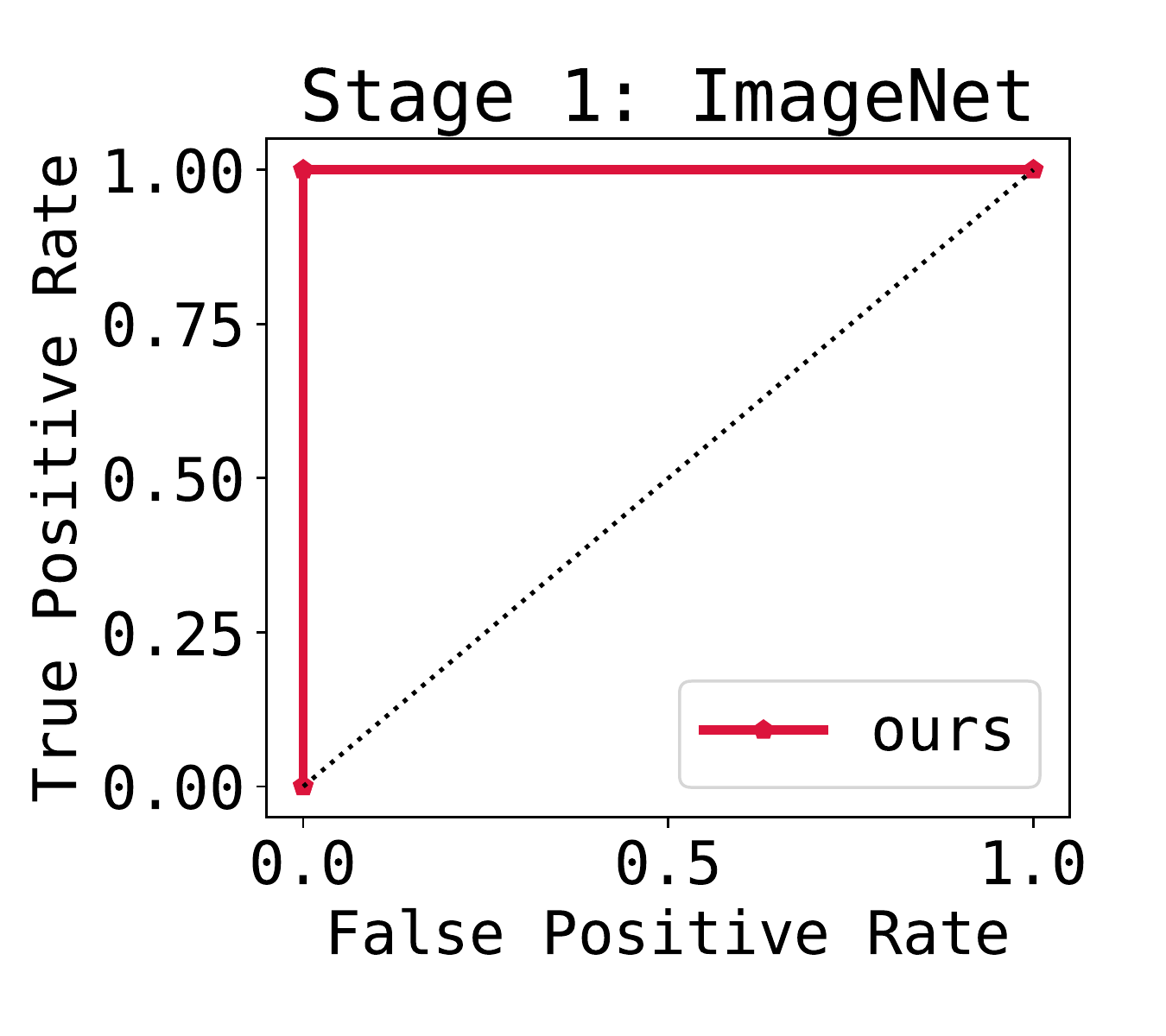}
\end{minipage}
%
%
\caption{The ROC curves of $F$ against $W$'s outputs on the FFHQ (left) and ImageNet (right) datasets in the first stage ($n=1$).}
\label{fig:stage-1-roc}
\vspace{-2.0em}
\end{wrapfigure}

\begin{figure}[t]
\centering
\includegraphics[width=.85\columnwidth]{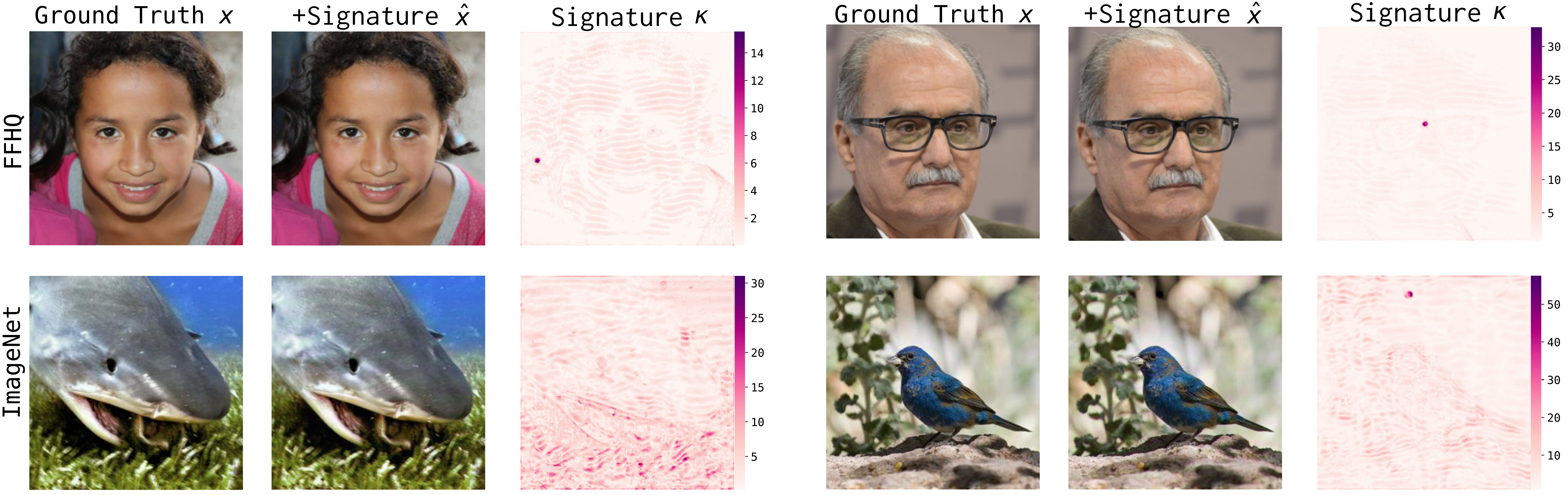}
\vskip-7pt\caption{Sample outputs from the signature injector $W$ in stage one.
The two columns on the left correspond to FFHQ, while the rest correspond to ImageNet.
The signature $\kappa$ is visualized as the pixel-wise L-$2$ norm, where the peak value varies across inputs.
}
\label{fig:vis}
\vspace{-2em}
\end{figure}


In the first stage, we validate the signature injector $W$ for the image quality, and the classifier $F$ for the accuracy against the outputs of $W$.
%
%
The experiments are conducted on FFHQ and ImageNet, respectively.
The corresponding results are shown in Table~\ref{tab:stage-1} and Table~\ref{tab:stage-1-n} for the $n=1$ case and the $n=2$ case, respectively.

According to Table~\ref{tab:stage-1}, when the binary code is $1$-bit,
the adversarial signature can be added to the outputs of $W$ while retaining good image quality.
This is reflected by  $51.4$ PSNR and $0.52$ FID on FFHQ, and  $38.4$ PSNR and $5.71$ FID on ImageNet.
The results on ImageNet (natural images) are slightly worse than that on FFHQ (face images) due to the more complex distribution. 
Some images with signature are visualized in Figure~\ref{fig:vis}.
%

Apart from the injector, the classifier $F$ achieves $100.0\%$ and $99.9\%$ accuracy on FFHQ and ImageNet, respectively.
The corresponding ROC curves can be found in Figure~\ref{fig:stage-1-roc}.
These results suggest that although the learned signatures are small in L-$2$ norm,
they are still highly recognizable by $F$.
%

\begin{figure}[t]
\centering
\includegraphics[width=0.7\columnwidth]{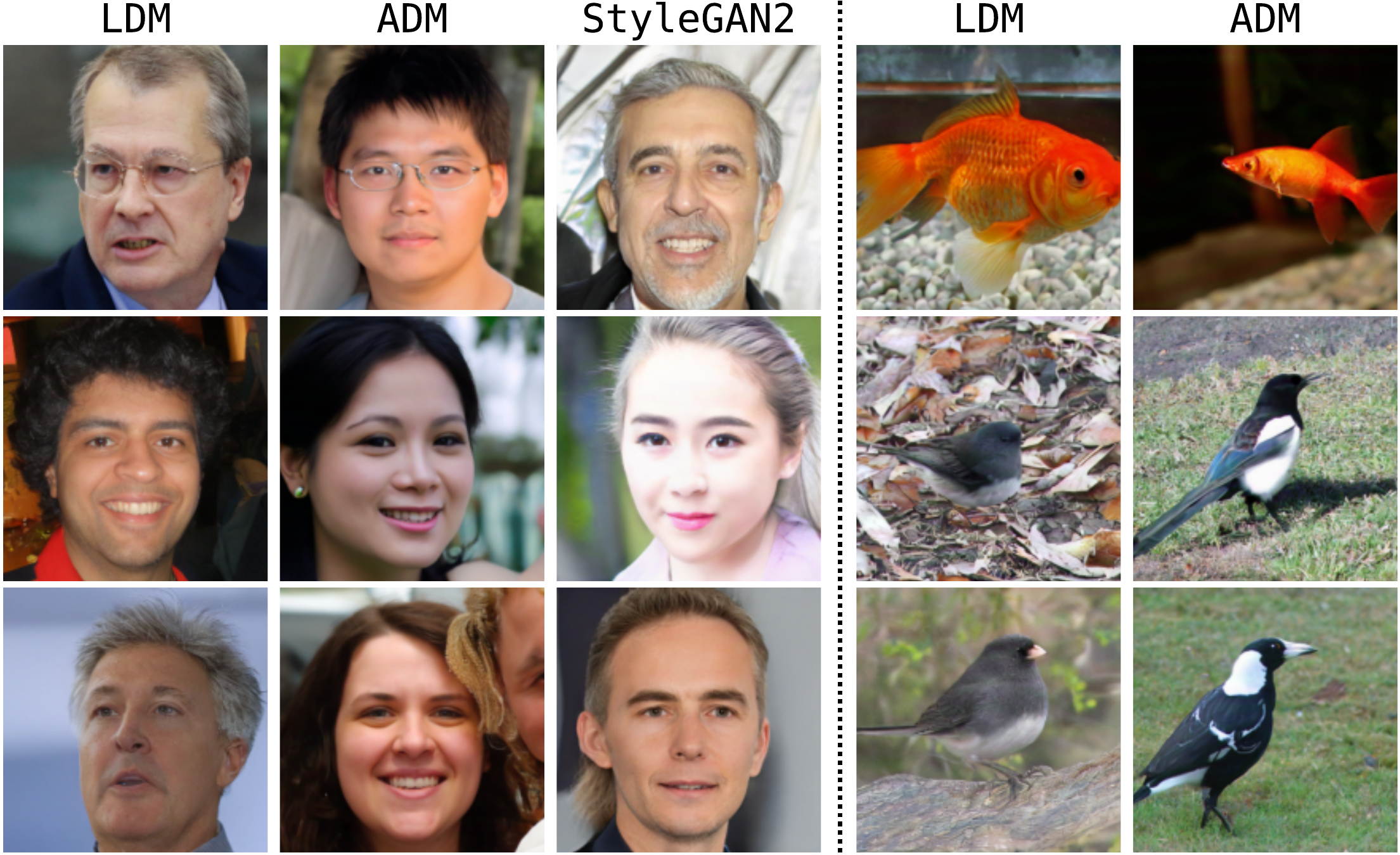}
\vskip-7pt\caption{Sample outputs from the fine-tuned generator $\hat{G}$ in stage two.
The three columns on the left correspond to the FFHQ dataset, while the two on the right correspond to the ImageNet dataset.
}
\label{fig:visg}
\vspace{-1.5em}
\end{figure}

According to Table~\ref{tab:stage-1-n}, 
when the binary code length is $n=2$, 
our method remains effective, as suggested by the good image quality for $W$ and high classification accuracy of $F$.
Notably, since the $n=2$ case requires $W$ to learn different variants of $\kappa$ for different binary codes, the learning becomes more difficult than in the $n=1$ case, resulting in a slight performance gap. 
%

%

\subsection{Validating $\hat{G}$ and $F$ in the Second Stage}
\label{sec:42}

\begin{table}[t]
\begin{minipage}{0.50\textwidth}
\caption{Validating $\hat{G}$ and $F$ in the second stage when the length of binary code is $n=1$. ``FID$^*$'' is for the official pre-trained model released by the corresponding authors;
``FID'' is for the model reproduced with the respective official code;
``Acc.'' is the generated/real image classification accuracy.
}
\label{tab:stage-2}
\centering%
 \resizebox{\columnwidth}{!}{%
\setlength{\tabcolsep}{4.5pt}
\begin{tabular}{cc|cc|c|c}
\toprule
\multirow{2}{*}{\bf Dataset} & \multirow{2}{*}{\bf Generator} & \multicolumn{2}{c|}{$G$} & \multicolumn{1}{c|}{$\hat{G}$} & $F$\tabularnewline
 &  & FID$^*$ $\downarrow$ & FID $\downarrow$ & FID $\downarrow$ & Acc. (\%) $\uparrow$\tabularnewline
 \midrule
\multirow{3}{*}{FFHQ} & LDM & 9.36 & 9.70 & 9.20 & 100.0\tabularnewline
 & ADM & - & 12.32 & 13.61 & 100.0\tabularnewline
 & StyleGAN2 & 4.16 & 3.97 & 4.35 & 99.9\tabularnewline
\midrule
\multirow{2}{*}{ImageNet} & LDM & 7.41 & 6.72 & 5.48 &100.0 \tabularnewline
 & ADM & 6.38 & 7.36 & 6.65 & 100.0\tabularnewline
 \bottomrule
\end{tabular}
 }
\end{minipage}
\hfill
%
\begin{minipage}{0.48\textwidth}
\caption{Validating $\hat{G}$ and $F$ in the second stage when the length of binary code is $n=2$. 
See the caption for Table~\ref{tab:stage-2} for the meaning of ``FID$^*$'', ``FID'', and ``Acc.''.
}
\label{tab:stage-2-n}
\centering%
 \resizebox{\columnwidth}{!}{%
\setlength{\tabcolsep}{4.5pt}
\begin{tabular}{cc|cc|c|c}
\toprule
\multirow{2}{*}{\bf Dataset} & \multirow{2}{*}{{\bf Model} ({\tt code})} & \multicolumn{2}{c|}{$G$} & \multicolumn{1}{c|}{$\hat{G}$} & $F$\tabularnewline
 &  & FID$^*$ $\downarrow$ & FID $\downarrow$ & FID $\downarrow$ & Acc. (\%) $\uparrow$\tabularnewline
 \midrule
\multirow{3}{*}{FFHQ} & LDM ({\tt 01}) & \multirow{3}{*}{\rotatebox{45}{9.36}} & \multirow{3}{*}{\rotatebox{45}{9.70}} & 9.86 &  \multirow{3}{*}{\rotatebox{45}{99.8}} \tabularnewline
 & LDM ({\tt 10}) &  & & 9.20 & \tabularnewline
 & LDM ({\tt 11}) &  &   & 10.35 & \tabularnewline
 \bottomrule
\end{tabular}
 }
\end{minipage}
\vspace{-1em}
\end{table}

\begin{figure}[t]
\centering
\begin{minipage}{0.28\columnwidth}
\includegraphics[width=\columnwidth]{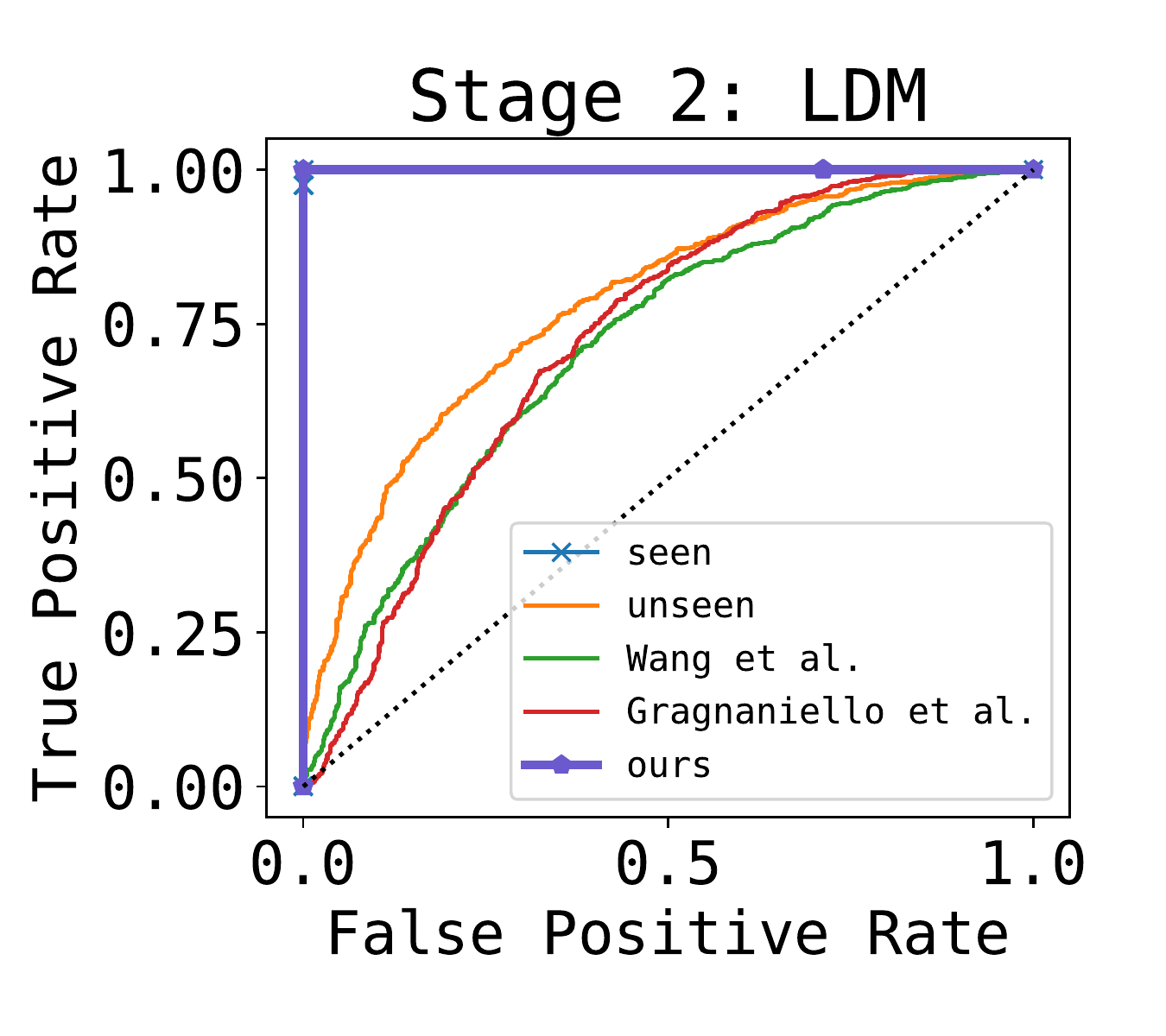}
\end{minipage}
~
\begin{minipage}{0.28\columnwidth}
\includegraphics[width=\columnwidth]{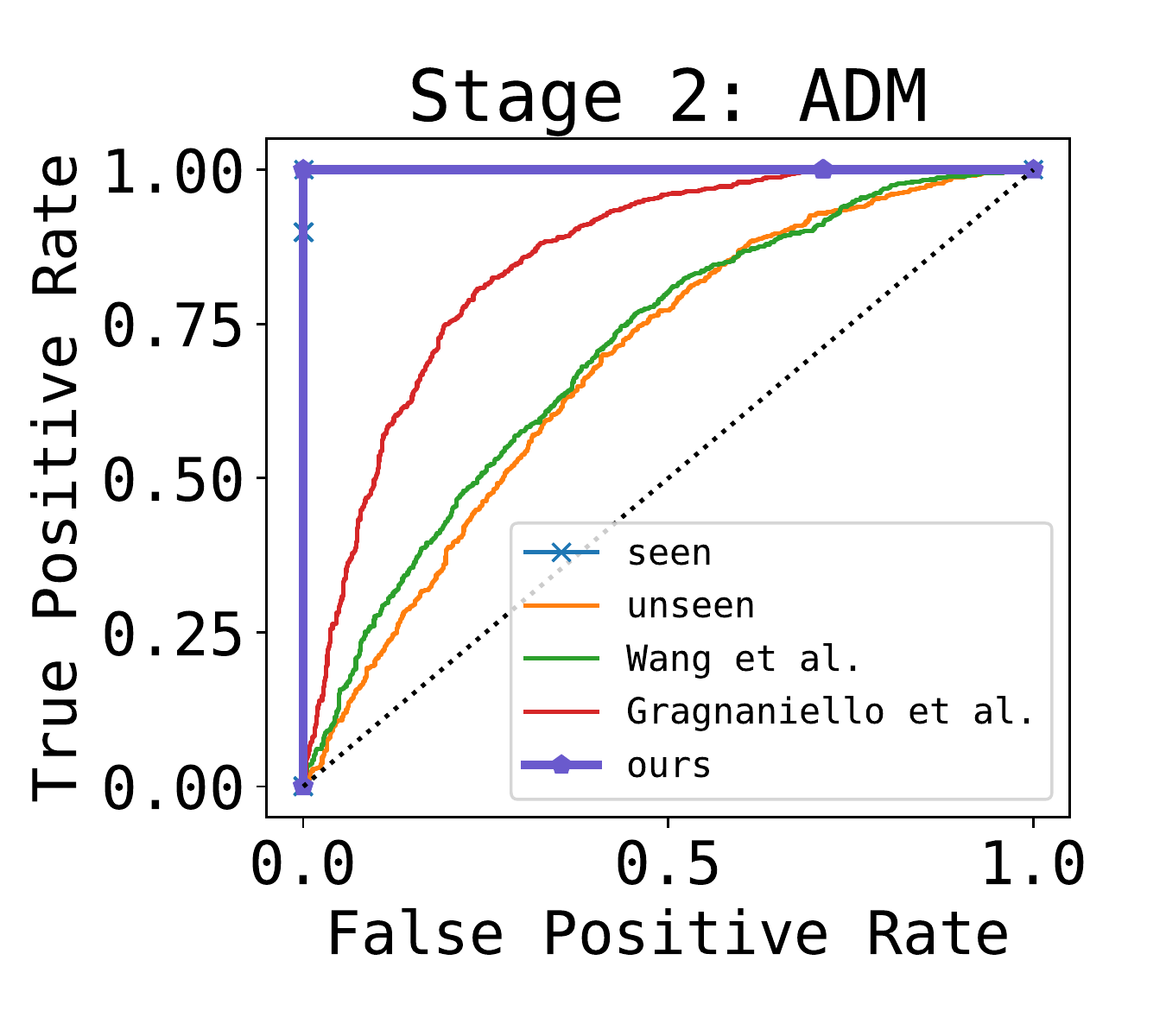}
\end{minipage}
~
\begin{minipage}{0.28\columnwidth}
\includegraphics[width=\columnwidth]{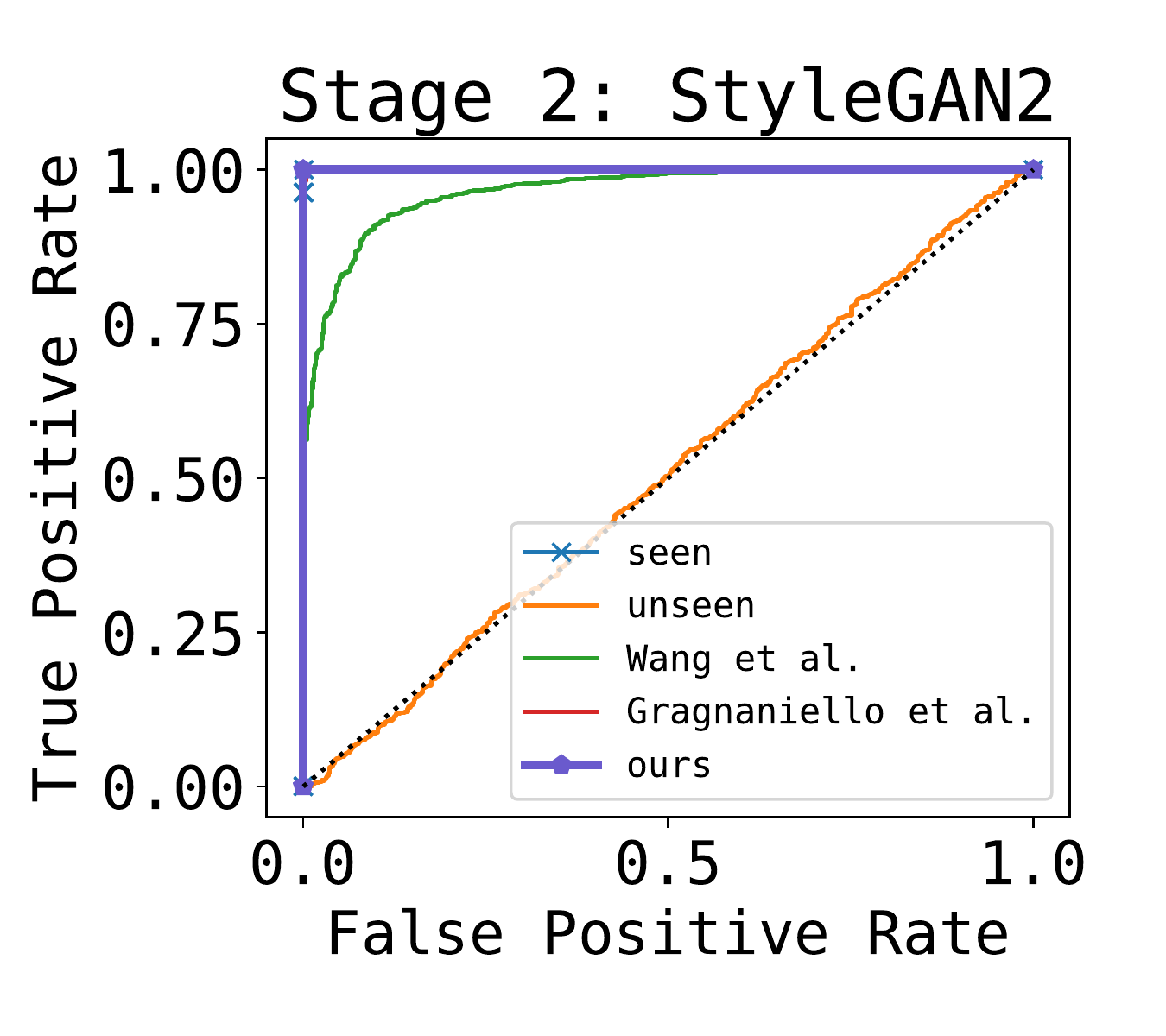}
\end{minipage}
\vspace{-1.0em}
\caption{The ROC curves of $F$ against $\hat{G}$ (for our method) and $G$ (for baseline method) in the second stage ($n=1$). The baseline method involves the ``seen'' case and the ``unseen'' case.}
\label{fig:stage-2-roc}
\vspace{-1em}
\end{figure}


In the second stage, a pre-trained $G$ is fine-tuned with $W$ and $F$ being fixed.
We conduct experiments accordingly to validate the fine-tuned generator $\hat{G}$, and the classifier $F$ against the outputs of $\hat{G}$.
The results can be found in Table~\ref{tab:stage-2} and Table~\ref{tab:stage-2-n}
for the $n{=}1$ and $n{=}2$ cases, respectively.

According to Table~\ref{tab:stage-2}, when the binary code length is $n=1$, the generator $\hat{G}$ can learn the adversarial signatures from $W$, which makes its outputs more recognizable by $F$.
Take the LDM model on the FFHQ dataset as an example.
The fine-tuned model $\hat{G}$ achieves a similar FID to its original counterpart $G$.
%
This indicates no significant output quality difference between $G$ and $\hat{G}$.
To demonstrate this qualitatively, we visualize some generated images in Figure~\ref{fig:visg}.
%

Although the adversarial signatures the generator $\hat{G}$ has ``inherited'' are imperceptible, they are still highly recognizable by $F$.
This is quantitatively demonstrated by the $100.0\%$ generated/real image classification accuracy.
The corresponding ROC curves can be found in Figure~\ref{fig:stage-2-roc}.
%

According to Table~\ref{tab:stage-2-n}, when the binary code length is $n=2$, the adversarial signatures can also be effectively learned by the generators, which can still be detected by $F$.
%

%


\subsection{Comparison to State-of-the-art Methods}
\label{sec:43}

\begin{wraptable}{r}{0.48\linewidth}
\vspace{-3em}%
\centering%
\caption{Generated image detection accuracy with 95\% error bars. 
The first four rows are based on the official pre-trained generators. The last row is based on fine-tuned generators. }
\label{tab:sota}%
\setlength{\tabcolsep}{9pt}%
 \resizebox{1.0\linewidth}{!}{
\begin{tabular}{c|c|c|c}
\toprule
Detector \textbackslash{} Generator & LDM & ADM & StyleGAN2 \tabularnewline
\midrule
Baseline (Seen) & 99.9$\pm$.006 &  99.7$\pm$.003 & 99.9$\pm$.006\tabularnewline
Baseline (Unseen) & 51.6$\pm$.031 & 49.8$\pm$.031 & 49.9$\pm$.031\tabularnewline
\cite{wang2019cnngenerated} & 50.4$\pm$.031 & 49.9$\pm$.031 & 66.7$\pm$.029 \tabularnewline
\cite{gragnaniello2021gan} & 50.2$\pm$.031 & 49.9$\pm$.031 & 97.8$\pm$.009  \tabularnewline
\hline
Ours & 100.0$\pm$.00 & 100.0$\pm$.00 & 99.9$\pm$.006 \tabularnewline
\bottomrule
\end{tabular}
 }
 \vspace{-1em}
\end{wraptable}

After verifying the effectiveness of our proposed method, we compare it with a baseline method and the state-of-the-art methods on FFHQ.

The baseline method corresponds to directly training the classifier $F$ (ResNet-34) to differentiate the generated images $\bm{y}$ from the original images  $\bm{x}$.
As shown in the first row of Table~\ref{tab:sota}, if all three generators (\emph{i.e.}, LDM, ADM, and StyleGAN2) are \emph{seen} by $F$, its accuracy is close to $100\%$.
However, in the second row, the baseline method suffers from poor generalization against \emph{unseen} generators under the leave-one-out setting.
For instance, in the first column, the ADM and StyleGAN2 are seen by $F$, but not LDM.
The accuracy of $F$ against the LDM outputs drops to mere $51.6\%$.
The corresponding ROC curves can be found in Figure~\ref{fig:stage-2-roc}.

The generalization issue against \emph{unseen} generators also exists with the state-of-the-art methods including \cite{wang2019cnngenerated,gragnaniello2021gan}, as shown in Table~\ref{tab:sota}.
In contrast, our method can reuse  the $W$ and $F$ for any generative model, and achieve high accuracy as long as its input is from a fine-tuned generator.


\begin{table}[t]
\begin{minipage}[t]{0.49\linewidth}
\vspace{-0.5em}
\centering
\caption{Model sensitivity to $\lambda$ on FFHQ dataset.}
\label{tab:lambda}
\setlength{\tabcolsep}{4pt}
 \resizebox{\columnwidth}{!}{%
\begin{tabular}{cc|cccccccc}
\toprule
\multicolumn{2}{c|}{$\lambda$} & 1.0 & 0.1 & 0.05 & 0.01 & 0.005 & 0.001 & 0.00001 & 0.0\tabularnewline
\midrule
\multirow{2}{*}{$W$} & PSNR $\uparrow$ & 37.5 & 42.8 & 51.4 & 52.1 & 52.7 & 54.9 & 63.0 & 67.5\tabularnewline
 & FID $\downarrow$ & 9.33 & 3.16 & 0.52 & 0.46 & 0.44 & 0.39 & 0.06 & 0.02\tabularnewline
 \hline
$F$ & Acc. (\%) $\uparrow$ & 100.0 & 100.0 & 100.0 & 100.0 & 100.0 & 100.0 & 99.6 & 50.0\tabularnewline
\bottomrule
\end{tabular}
 }
 \end{minipage}
%
\hfill
\begin{minipage}[t]{0.49\linewidth}
\vspace{-0.5em}
\caption{Robustness of adversarial signature after common image transformations.}
\label{tab:transformation}
\centering
 \resizebox{\columnwidth}{!}{
\begin{tabular}{c|ccc}
\toprule
Transformation & Gaussian blur & Horizontal flip & Rotation \tabularnewline
\midrule
$F(\hat{G}(\bm{z}))$ Acc. (\%) $\uparrow$ & 100.0 &  100.0& 100.0\tabularnewline
\bottomrule
\end{tabular}
 }
\end{minipage}

%
\begin{minipage}[t]{0.49\linewidth}
\caption{Validating $W$ and $F$ when they are separately trained.
The length of the binary code is $n=1$.
The dataset is FFHQ.
}
\label{tab:separately}
\begin{center}
 \resizebox{1.0\columnwidth}{!}{%
\setlength{\tabcolsep}{7pt}
\begin{tabular}{c|cc|c}
\toprule
\multirow{2}{*}{Model} & \multicolumn{2}{c|}{Signature Injector $W$} & Classifier $F$\tabularnewline
 & PSNR $\uparrow$ & FID $\downarrow$ & Accuracy (\%) $\uparrow$\tabularnewline
\midrule
Jointly & 51.4 & 0.52 & 100 \tabularnewline
Separately & 23.4& 140 & 50.1 \tabularnewline
\bottomrule
\end{tabular}
 }
\end{center}
\end{minipage}
\hfill
\begin{minipage}[t]{0.49\linewidth}
\caption{Validating $W$, $F$, and $M$ in the first stage with $1$-bit binary code.
The noise is Gaussian with zero mean and $0.5$ standard deviation.
%
%
}
\label{tab:m}
\begin{center}
 \resizebox{1.0\columnwidth}{!}{%
\setlength{\tabcolsep}{5pt}
\begin{tabular}{c|c|c|c}
\toprule
\multirow{2}{*}{Setting} & $W$ & $F(W(\bm{x}))$ & $F(M(W(\bm{x})))$ \tabularnewline
 & PSNR $\uparrow$ & Acc. (\%) $\uparrow$ & Acc. (\%) $\uparrow$
 \tabularnewline
\midrule
(w/o noise $\bm{e}$, w/o $L_\text{aux}$) &51.4 &100.0& 50.0 \tabularnewline
(w/o noise $\bm{e}$, w/ $L_\text{aux}$)~~ & 44.8 & 99.9 & 52.6\tabularnewline
~~(w/ noise $\bm{e}$, w/o $L_\text{aux}$) &29.4 &99.9 &50.0 \tabularnewline
(w/ noise $\bm{e}$, w/ $L_\text{aux}$) &29.1 &99.9 & 98.2 \tabularnewline
\bottomrule
\end{tabular}
 }
\end{center}
\end{minipage}

\vspace{-1em}
\end{table}


\section{Discussions}
\label{sec:5}

In this section, we study the sensitivity of $\lambda$ in Eq.\eqref{eq:advwm}, and some alternative method designs.
We also discuss how the desired characteristics mentioned in  Section~\ref{sec:1} are satisfied.

\subsection{Parameter Sensitivity of $\lambda$ \& Pre-trained $F$}

\noindent\textbf{Sensitivity of $\lambda$.}
In Eq.~\eqref{eq:advwm}, the parameter $\lambda$ balances the two loss terms $L_\text{rec}$ and $L_\text{cls}$, which are adversarial against each other as discussed in Section~\ref{sec:32}.
%
%
We conduct experiments with varying $\lambda$ values on FFHQ for the first stage, in order to study the sensitivity of $\lambda$.
The results are shown in Table~\ref{tab:lambda}.
When $\lambda$ is gradually decreased from $1.0$,
the accuracy of $F$ is not very sensitive.
However, a clear trend can be seen where $W$ tries to sacrifice image quality in exchange for a lower cross-entropy loss.
When $\lambda=0$, $W$ is expected to learn the identity mapping, and $F$ is not trained.
As a result, the reconstructed image is of high quality, and $F$ behaves the same as a random classifier.
Most importantly, a nearly optimal pair of $W$ and $F$ can be found even if $\lambda$ is very small, which leads to a negligible image quality drop.
This supports our theory in Proposition~\ref{prop1}.


\noindent\textbf{Pre-trained $F$.}
To better understand the distinction between adversarial signatures and the features used by baseline detectors, we replace the $F$ with the pre-trained and fixed ``Baseline (Seen)'' classifier from Table~\ref{tab:sota} in the first stage.
This leads to significantly worse performance as shown in Table~\ref{tab:separately}. 
The results suggest that there is hardly any resemblance in features between our signature-based classifier and a baseline classifier.
%
%
Therefore, adversarial signature is different from the features used by the baseline detectors, and $W$ and $F$ should be jointly optimized.
%


\begin{figure}[H]
\includegraphics[width=.9\columnwidth]{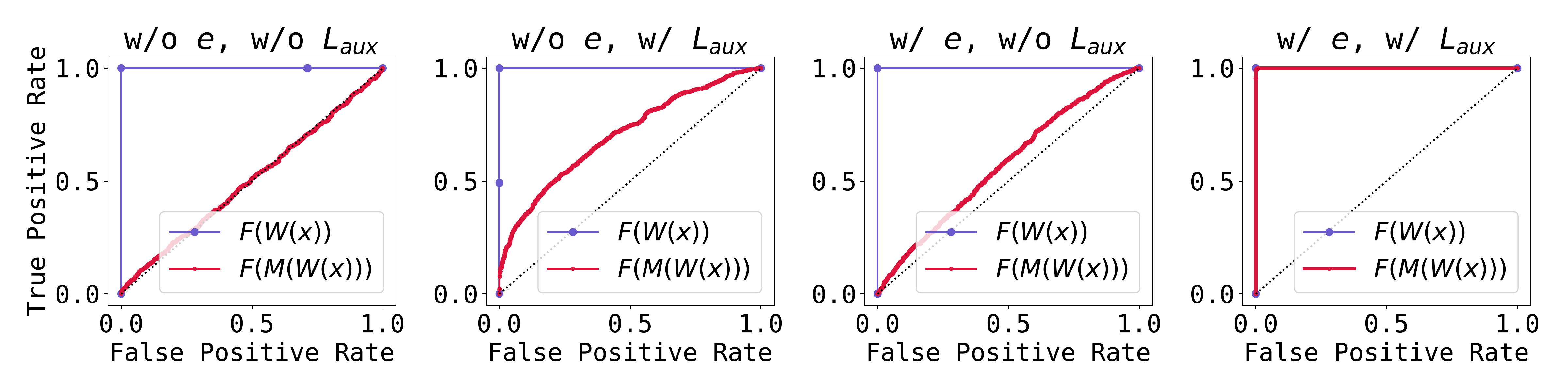}\\
\vspace{-2em}
\caption{ROC for $F(W(\bm{x}))$ \& $F(M(W(\bm{x})))$ in Table~\ref{tab:m}.}
\label{fig:fm}
\vspace{-1em}
\end{figure}

\subsection{Characteristics of Adversarial Signature}
\label{sec:discuss-char}

\textbf{Imperceptibility.}
%
%
This is enforced by Eq.~\eqref{eq:rec}.
%
The imperceptibility is demonstrated by
Table~\ref{tab:stage-1}-\ref{tab:stage-1-n}, Figure~\ref{fig:vis} for the outputs of $W$;
and Table~\ref{tab:stage-2}-\ref{tab:stage-2-n}, Figure~\ref{fig:visg} for the outputs of $\hat{G}$.


\textbf{Persistency.} 
\textbf{(1)} To make the signature in $\hat{\bm{y}}$ hard to be invalidated by image transformations, 
it is learned with data augmentation (see Section~\ref{sec:4}).
According to Table~\ref{tab:transformation}, $F$ has gained robustness against the image transformations.
\textbf{(2)} A possible adaptive attack from a malicious user may involve obtaining the inverse function of $W$, namely the restoration attack mentioned in Proposition~\ref{prop3}.
To achieve this, $M$ learns to restore the original image $\bm{x}$ from the signed image $\hat{\bm{x}}$:
%
$L_\text{M} =
\| M[W(\bm{x})] - \bm{x} \|^2.
$
Accordingly, the classifier $F$ has to recognize the outputs of $M$ by an extra loss term on top of Eq.~\eqref{eq:advwm}:
$L_\text{aux} {=} \mathbb{E}_{M}\{\log (1 - F(M[W(\bm{x})]))\}$.
%
In the implementation, we approximate the expectation over $M$ using multiple snapshots of $M$ jointly trained with $W,F$.
The experimental results on FFHQ can be found in Table~\ref{tab:m} and Fig.~\ref{fig:fm}. 
The default setting (Table~\ref{tab:stage-1}) is without the noise $\bm{e}$ (see Section~\ref{sec:31}), nor the $L_\text{aux}$. 
%
When both the noise $\bm{e}$ and $L_\text{aux}$ are applied,
it is still difficult to remove the adversarial signatures even if the proposed method is disclosed to malicious users.
%
The results support  Proposition~\ref{prop3}.

\textbf{Inevitability.}
Once the generative model is fine-tuned, the adversarial signature will be inevitably included in its outputs.
%
%
Restoring $G$ from $\hat{G}$ may require access to the training images without signatures, with which a malicious user can already train new generators instead of using $\hat{G}$.


\textbf{Efficiency.}
(1) Inference:
Our method only changes the generative model parameters.
The inference cost for $\hat{G}$ is identical to that of $G$.
(2) Training:
Assume $r$ generative models are to be released one by one.
The complexity of re-training a detector every time upon the release of a new generator is $O(r^2)$.
%
In contrast, the complexity of the proposed method is $O(r)$, 
because $W$ and $F$ are reused once trained.
%
%
Our method is efficient in terms of complexity.

\textbf{Limitations.}
(1) The binary code length $n$ limits the amount of additional information it can represent.
%
(2) We assume the training dataset without adversarial signature is not available to malicious users.
But once it is available, the malicious user is able to train a new generator instead of using $\hat{G}$.
%
%

\section{Conclusions}

The proposed method aims to modify a given generative model, making its outputs more recognizable due to adversarial signatures.
The adversarial signature can also carry additional information for tracking the source of generated images.
The experimental results on two datasets demonstrate the effectiveness of the proposed method.


\cleardoublepage
\setcounter{page}{1}
\setcounter{section}{0}
\setcounter{figure}{0}
\setcounter{table}{0}
\renewcommand\thesection{\Alph{section}} 
\renewcommand\thefigure{\Alph{figure}} 
\renewcommand\thetable{\Alph{table}} 

\section*{\centering Supplementary Material}
\section{Additional Results}
\begin{figure}[h]
\begin{center}
\includegraphics[width=0.9\textwidth]{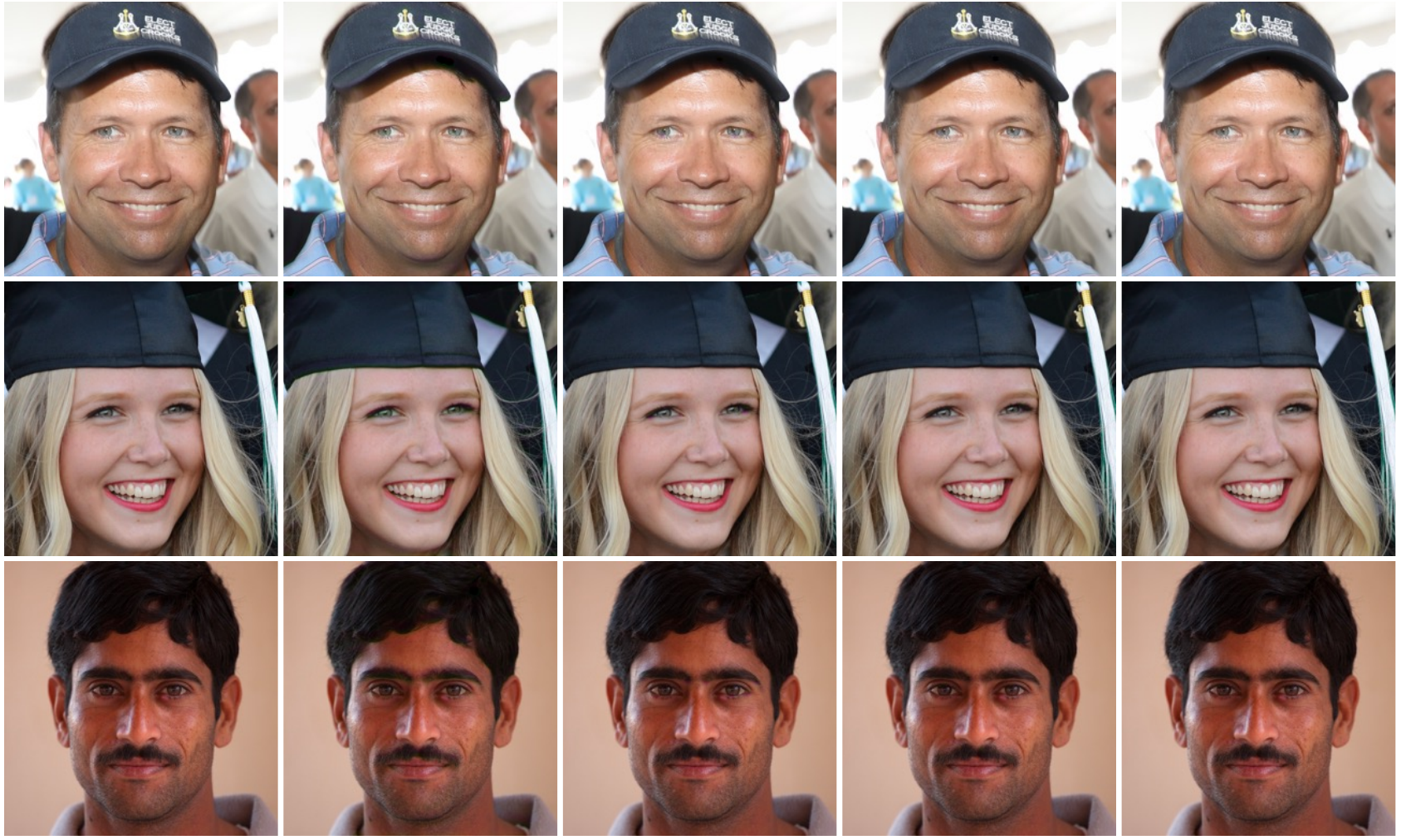}
\end{center}
\caption{Visualization of the signed images with varying $\lambda$.
}
\label{fig:supp1}
\end{figure}

Table~6 in the main paper shows the effect of varying the parameter $\lambda$ on the PSNR, FID and classification accuracy. Here we visualize the signed images with different $\lambda$ in Fig~\ref{fig:supp1}. We can see that the signed images are almost visually indistinguishable from the original images for $\lambda \in [1e-5,0.1]$.  

\section{Proof of Propositions}
\begin{proposition}
\label{prop1}
(\textit{\textbf{Imperceptibility}}) There exist optimal pairs of signature injector $W$ and classifier $F{:}~ \mathbb{R}^n {\mapsto} \{0,1\}$, so that for any image $\forall \bm{x} {\in} \mathbb{R}^n$, $\forall \epsilon {>}0$,
its distance to the signed image $W(\bm{x})$
is smaller than $\epsilon$, 
and $F$ correctly discriminates them, 
\emph{i.e.},
$\norm{W(\bm{x}){-}\bm{x}}{<}\epsilon$, and
$F(W(\bm{x}))\ne F(\bm{x})$. 
\end{proposition}
\noindent\textit{Proof.}
For simplicity, we consider the case when $x\in \mathbb{R}^1$.
Let $W(x)$ be an arbitrary irrational number within $(x-\epsilon, x+\epsilon)$ when $x$ is rational, and otherwise an arbitrary rational number within $(x-\epsilon, x+\epsilon)$.
Let $F$ be a classifier that discriminates rational/irrational numbers.
This pair of $W, F$ satisfies the given condition, and proves the existence of optimal watermarking systems. 
\begin{remark}
The $W, F$ presented in the proof are not feasible for implementation in practice.
However, when $W, F$ are deep neural networks, the existence of adversarial samples~\cite{l-bfgs} implies that 
one can find a $W(\bm{x})$ that flips the prediction of $F$ while being very close to $\bm{x}$.
\end{remark}

\begin{proposition}
\label{prop2}
Let $\bm{e}$ be a zero-mean, positive-variance additive noise. There exist noise augmented $W, F$ that satisfy the following condition: $\forall \epsilon > 0, \mathbb{E}_{\bm{e}}[\norm{W(\bm{x}+\bm{e})-\bm{x}}]< \epsilon$ and $F(W(\bm{x}))\ne F(\bm{x})$. 
\end{proposition}
\noindent \textit{Proof.}
We can prove the existence of such $W, F$ by constructing an example similar to the one in Proposition~\ref{prop1} and set $\bm{e}$ to a rational noise.
The existence of such $W, F$ can be proved in a similar way as Proposition~\ref{prop1}, by setting $\bm{e}$ to a rational noise.
Then we have $\mathbb{E}[ \norm{W(\bm{x}+\bm{e})-\bm{x}} ] = \mathbb{E}[ \norm{W(\bm{x}+\bm{e})-(\bm{x}+\bm{e}) + \bm{e}} ] \le \mathbb{E}[ \norm{W(\bm{x}+\bm{e})-(\bm{x}+\bm{e})}] + \mathbb{E}[\norm{\bm{e}}] < \epsilon$. 

\begin{lemma}
\label{prop00}
Let $\bm{x}$ and $\bm{e}$ be zero-mean positive-variance random variables. 
For any non-constant mapping $M$, we have ~ $\mathbb{E}_{\bm{x},\bm{e}}[\norm{ M(W(\bm{x}+\bm{e})) -\bm{x}} ] > 0$.
\end{lemma}
\noindent\textit{Proof.} Assume that $\mathbb{E}[\norm{ M(W(\bm{x}+\bm{e})) -\bm{x}} ] = 0$. Then $\forall \bm{x}, \bm{e}$, $M(W(\bm{x}+\bm{e}))=\bm{x} $. If we let $\bm{x}=\bm{0}$, then $M(W(\bm{e}))=\bm{0}$, which is contradictory to the definition of $M$. 
Since the equal sign does not hold, and an L-$2$ norm is always greater than or equal to $0$, we have
$\mathbb{E}[\|M(W(\bm{x}+\bm{e})) - \bm{x}\|] > 0$.

\begin{proposition}
\label{prop3}
(\textit{\textbf{Persistency}}) The noise augmented
$W, F$ stated in Proposition~\ref{prop2} is robust to image restoration attack, as optimizing $\min_M \mathbb{E}_{\bm{x},\bm{e}}[ \norm{ M(W(\bm{x}+\bm{e})) - \bm{x} }]$ will result in $M$ being an identity mapping. 
\end{proposition} 
\noindent \textit{Proof. }Please refer to the supplementary material. 
\emph{Proof.} As shown in Proposition~\ref{prop2}, $\forall \epsilon > 0$,  $\mathbb{E}[\norm{W(\bm{x}+\bm{e})-\bm{x}}]< \epsilon$.
According to Lemma~\ref{prop00}, we have $\mathbb{E}[\norm{ M(W(\bm{x}+\bm{e})) -\bm{x}}]>0$.
Therefore, for any mapping $M$, $\mathbb{E}[\norm{W(\bm{x}+\bm{e})-\bm{x}}]\le\mathbb{E}[\norm{ M(W(\bm{x}+\bm{e})) -\bm{x}}]$.
Hence, $W(M(\bm{x}))=M(\bm{x})$ is the solution for $\min_M \mathbb{E}[ \norm{ M(W(\bm{x}+\bm{e})) - \bm{x} }]$.

\section{Broader Impact}
This work is intended to develop a system to mitigate the risk of image generation models by tracking the source of generated images based on signatures. Malicious users may attack this system with fake signatures, \textit{e.g.} by adding a signature on a real image to make it classified as generated, and compromise the credibility of true information. Potential mitigation strategies includes gated release of the watermark injectors, the use of longer multi-bit code and only releasing the code to the corresponding owners of generative models. 

\section{Limitations}
A limitation of the proposed method is that it requires to finetune a pretrained generative model to embed the signature. A direction for future work is to explore the training-free framework to secure deep generative models,~\textit{e.g.} by directly modifying model parameters. 

\section{Compute}
The signature injector is trained on a RTX A6000 GPU. The generative models are finetuned using 4 RTX A6000 GPUs.

\bibliography{neurips_2023}
\bibliographystyle{unsrt}

\end{document}